\documentclass[conference]{IEEEtran}
\IEEEoverridecommandlockouts
\usepackage{cite}
\usepackage{amsmath,amssymb,amsfonts}
\usepackage{algorithmic}
\usepackage{graphicx}
\usepackage{textcomp}
\usepackage{xcolor}
\usepackage[ruled,linesnumbered]{algorithm2e}
\usepackage{multirow}
\usepackage{enumerate}
\usepackage{url}
\usepackage{booktabs}
\usepackage{threeparttable}
\usepackage{endnotes}
\usepackage{bbding}
\usepackage{bm}
\usepackage{CJKutf8}
\usepackage[utf8]{inputenc}
\usepackage{subfigure}

\def\BibTeX{{\rm B\kern-.05em{\sc i\kern-.025em b}\kern-.08em
    T\kern-.1667em\lower.7ex\hbox{E}\kern-.125emX}}

\begin{document}

\title{
Fast and Accurate Recognition of Chinese Clinical Named Entities with Residual Dilated Convolutions
}

\author{\IEEEauthorblockN{Jiahui Qiu$^1$, Qi Wang$^1$, Yangming Zhou$^{1,*}$, Tong Ruan$^{1,*}$ and Ju Gao$^2$}
\IEEEauthorblockA{$^1$School of Information Science and Engineering, East China University of Science and Technology, Shanghai 200237, China \\
$^2$Shanghai Shuguang Hospital, Shanghai 200120, China \\
$^*$Corresponding authors\\
Emails: \{ymzhou,ruantong\}@ecust.edu.cn\\
}
}

\maketitle

\begin{abstract}

Clinical Named Entity Recognition (CNER) aims to identify and classify clinical terms such as diseases, symptoms, treatments, exams, and body parts in electronic health records, which is a fundamental and crucial task for clinical and translation research. In recent years, deep learning methods have achieved significant success in CNER tasks. However, these methods depend greatly on Recurrent Neural Networks (RNNs), which maintain a vector of hidden activations that are propagated through time, thus causing too much time to train models. In this paper, we propose a Residual Dilated Convolutional Neural Network with Conditional Random Field (RD-CNN-CRF) to solve it. Specifically, Chinese characters and dictionary features are first projected into dense vector representations, then they are fed into the residual dilated convolutional neural network to capture contextual features. Finally, a conditional random field is employed to capture dependencies between neighboring tags. Computational results on the CCKS-2017 Task 2 benchmark dataset show that our proposed RD-CNN-CRF method competes favorably with state-of-the-art RNN-based methods both in terms of computational performance and training time.

\end{abstract}


\section{Introduction}
\label{Sec:Introduction}

Clinical Named Entity Recognition (CNER) is a critical task for extracting patient information from Electronic Health Records (EHRs) in clinical and translational research. CNER aims to identify and classify clinical terms in EHRs, such as diseases, symptoms, treatments, exams, and body parts. It is important to extract named entities from clinical texts because the clinical texts usually contains abundant healthcare information, while biomedical systems that rely on structured data are unable to access directly such information locked in the clinical texts. Identification of the clinical named entities is a non-trivial task. There are two main reasons. The one is the richness of EHRs, i.e., the same word or sentence can refer to more than one kind of named entities, and various forms can describe the same named entities~\cite{Gridach2017Character}. The other one is that a huge number of entities that rarely or even do not occur in the training set because of the use of non-standard abbreviations or acronyms, and multiple variations of same entities~\cite{sahu2017unified}. Furthermore, CNER in Chinese texts is more difficult compared to those in Romance languages due to the lack of word boundaries in Chinese and the complexity of Chinese composition forms~\cite{Duan2011A}.

Traditionally, rule-based approaches~\cite{Friedman1994A,Fukuda1998Toward}, dictionary-based approaches~\cite{rindflesch1999edgar,gaizauskas2000term} and machine learning approaches~\cite{Mccallum2000Maximum, Zhou2002Named, Mccallum2003Early} are applied to address the CNER tasks. Recently, along with the development of deep learning, some Recurrent Neural Network (RNN) based models, especially for the Bi-LSTM-CRF models~\cite{Gridach2017Character,Habibi2017Deep,wang2018incorporating}, have been successfully used and achieved the state-of-the-art results. However, RNN models are dedicated sequence models which maintain a vector of hidden activations that are propagated through time, thus requiring too much time for training.

To solve this problem, in this paper, we propose a Residual Dilated Convolutional Neural Network with Conditional Random Field (RD-CNN-CRF) for the Chinese CNER. In our method, Chinese CNER task is regarded as a sequence labeling task in character level in order to avoid introducing noise caused by segmentation errors, and dictionary features are utilized to help recognize rare and unseen clinical named entities. More specifically, Chinese characters and dictionary features are first projected into dense vector representations, then they are fed into the residual dilated convolutional neural network to capture contextual features. Finally, a conditional random field is employed to capture dependencies between neighboring tags. Computational studies on the CCKS-2017 Task 2 benchmark dataset\footnote{It is publicly available at \url{http://www.ccks2017.com/en/index.php/sharedtask/}} show that our proposed method achieves the highly competitive performance compared with state-of-the-art RNN-based methods, and is able to significantly save training time. In addition, we also observe that the Chinese CNER task do not necessarily rely on long-distance contextual information.

The main contributions of this work can be summarized as follows.
\begin{itemize}
    \item We propose a Residual Dilated Convolutional Neural Network with Conditional Random Field (RD-CNN-CRF) for the Chinese CNER. It is the first time to introduce the residual dilated convolutions for the CNER tasks especially for the Chinese CNER.
    \item Experimental results on the CCKS-2017 Task 2 benchmark dataset demonstrate that our proposed RD-CNN-CRF method achieves a highly competitive performance compared with state-of-the-art RNN-based methods. Moreover, our RD-CNN-CRF method is able to speed up the training process and save computational time.
\end{itemize}

The rest of the paper is organized as follows. We briefly review the related work on CNER and introduce the Chinese CNER in Section \ref{Sec:Related Work} and Section \ref{Sec:Chinses CNER}, respectively. In Section~\ref{Sec:RD-CNN-CRF Model}, we present the proposed RD-CNN-CRF model. We report the computational results in Section~\ref{Sec:Experimental Studies}. Section~\ref{Sec:Discussion and Analysis} is dedicated to experimentally investigate several key issues of our proposed model. Finally, conclusions are given in Section~\ref{Sec:Conclusion and Future Work}.

\section{Related Work}
\label{Sec:Related Work}

Due to the practical significance, Clinical Named Entity Recognition (CNER) has attracted considerable attention, and a lot of solution approaches have been proposed in the literature. All these existing approaches can be roughly divided into four categories: rule-based approaches, dictionary-based approaches, machine learning approaches and deep learning approaches.

Rule-based approaches rely on heuristics and handcrafted rules to identify entities~\cite{Friedman1994A,Zeng2006Extracting,Savova2010Mayo}. They were the dominant approaches in the early CNER systems. However, it is quite impossible to list all the rules to model the structure of clinical named entities, especially for various medical entities, and this kind of handcrafted approach always leads to a relatively high system engineering cost.

Dictionary-based approaches employ existing clinical vocabularies to identify entities~\cite{rindflesch1999edgar,gaizauskas2000term,song2015developing}. They were widely used because of their simplicity and their performance. A dictionary-based CNER system can extract all the matched entities defined in a dictionary from given clinical texts. However, it's unable to deal with out-of-dictionary entities, and consequently this kind of approach typically causes low recalls.

Machine learning approaches consider CNER as a sequence labeling problem where the goal is to find the best label sequence for a given input sentence~\cite{Lei2014A,lei2014named}. Typical methods are Hidden Markov Models (HMMs)~\cite{Zhou2002Named,song2015developing}, Maximum Entropy Markov Models (MEMMs) \cite{Mccallum2000Maximum,Finkel2004Exploiting}, Conditional Random Fields (CRFs)~\cite{Mccallum2003Early,Skeppstedt2014Automatic}, and Support Vector Machines (SVMs)~\cite{Wu2006Extracting,Ju2011Named}. However, these statistical methods rely on pre-defined features, which makes their development costly. More specifically, feature engineering process will cost much to find the best set of features which help to discern entities of a specific type from others. And it's more of an art than a science, incurring extensive trial-and-error experiments.

Deep learning approaches~\cite{Wu2015Named}, especially the methods based on Bidirectional RNN with CRF layer as the output interface (Bi-RNN-CRF)~\cite{huang2015bidirectional}, achieve state-of-the-art performance in CNER tasks and outperform the traditional statistical models~\cite{Gridach2017Character,Habibi2017Deep,Zeng2017LSTM}. RNNs with gated recurrent cells, such as Long-Short Term Memory (LSTM)~\cite{hochreiter1997long} and Gated Recurrent Units (GRU)~\cite{Cho2014Learning}, are capable of capturing long dependencies and retrieving rich global information. The sequential CRF on top of the recurrent layers ensures that the optimal sequence of tags over the entire sentence is obtained. Some scholars also tried to integrate other features like n-gram features~\cite{Ouyang2017Exploring} to improve the performance. However, RNNs are dedicated sequence models which maintain a vector of hidden activation that are propagated through time, so the RNN-based models often take long time for training.

\section{Chinese Clinical Named Entity Recognition}
\label{Sec:Chinses CNER}

The Chinese Clinical Named Entity Recognition (Chinese CNER) task can be regarded as a sequence labeling task. Due to the ambiguity in the boundary of Chinese words, following our previous work~\cite{Xia2017Clinical}, we label the sequence in the character level to avoid introducing noise caused by segmentation errors. Thus, given a clinical sentence $X=<x_1,...,x_n>$, our goal is to label each character $x_i$ in the sentence $X$ with BIEOS (Begin, Inside, End, Outside, Single) tag scheme. An example of the tag sequence for ``\begin{CJK*}{UTF8}{gbsn}腹平坦，未见腹壁静脉曲张。\end{CJK*}'' (The abdomen is flat and no varicose veins can be seen on the abdominal wall) can be found in Table \ref{table:labelingFormat}.

\begin{table*}[!ht]
\begin{center}
\caption{An Illustrative Example of Dictionary Features and Tags}
\label{table:labelingFormat}
\renewcommand\arraystretch{1.3}
\addtolength{\tabcolsep}{2pt}
\begin{tabular}{|c|c|c|c|c|c|c|c|c|c|c|c|c|c|}
\hline
Character Sequence  & \begin{CJK*}{UTF8}{gbsn}腹\end{CJK*}   & \begin{CJK*}{UTF8}{gbsn}平\end{CJK*} & \begin{CJK*}{UTF8}{gbsn}坦\end{CJK*} & \begin{CJK*}{UTF8}{gbsn}，\end{CJK*} & \begin{CJK*}{UTF8}{gbsn}未\end{CJK*} & \begin{CJK*}{UTF8}{gbsn}见\end{CJK*} & \begin{CJK*}{UTF8}{gbsn}腹\end{CJK*}   & \begin{CJK*}{UTF8}{gbsn}壁\end{CJK*}   & \begin{CJK*}{UTF8}{gbsn}静\end{CJK*}   & \begin{CJK*}{UTF8}{gbsn}脉\end{CJK*}   & \begin{CJK*}{UTF8}{gbsn}曲\end{CJK*}   & \begin{CJK*}{UTF8}{gbsn}张\end{CJK*}   & \begin{CJK*}{UTF8}{gbsn}。\end{CJK*} \\
\hline
Dict Feature Sequence & S-b & None & None & None & None & None & None & None & B-s & I-s & I-s & E-s & None \\
\hline
Tag Sequence & S-b & O & O & O & O & O & B-b & E-b & B-s & I-s & I-s & E-s & O \\
\hline
Entity Type & body & \multicolumn{5}{c|}{} & \multicolumn{2}{c|}{body} & \multicolumn{4}{c|}{symptom} &  \\
\hline
\multicolumn{14}{l}{\begin{tabular}[c]{@{}l@{}}\begin{minipage}{16.5cm}\vspace{1mm}\tiny \item[*] The B-tag indicates the beginning of an entity. The I-tag indicates the inside of an entity. The E-tag indicates the end of an entity. The O-tag indicates the character is outside an entity. The S-tag indicates the character is merely a single-character entity. As for entity types, the b-tag indicates the entity is a body part, and the s-tag indicates the entity is a symptom. Note that in this case the entity ``\begin{CJK*}{UTF8}{gbsn}腹壁\end{CJK*}'' is not included in the dictionary.\end{minipage} \end{tabular}}
\end{tabular}
\begin{tablenotes}
\tiny
\item[$\star$]
\end{tablenotes}
\end{center}
\end{table*}

\section{RD-CNN-CRF Model for the Chinese CNER}
\label{Sec:RD-CNN-CRF Model}

\begin{figure}[!htbp]
\begin{center}
\includegraphics[width=3.2in]{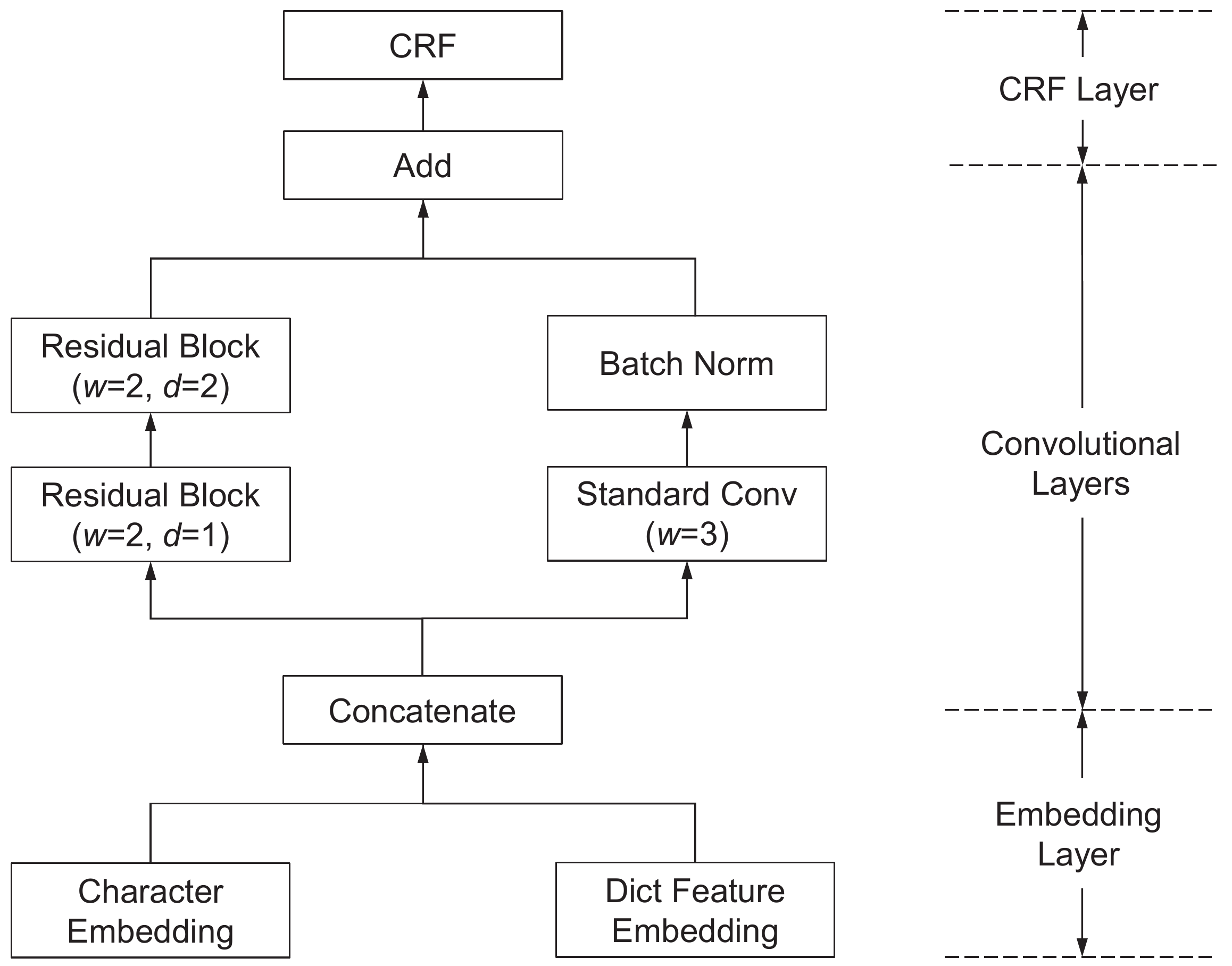}
\caption{Main architecture of our proposed RD-CNN-CRF Model.}
\label{fig:RDCNN-CRF}
\end{center}
\end{figure}

In this section, we present a Residual Dilated Convolutional Neural Network with Conditional Random Field (RD-CNN-CRF) for the Chinese CNER. As shown in Fig.~\ref{fig:RDCNN-CRF}, our proposed RD-CNN-CRF model consists of three key components: an embedding layer, some convolutional layers and a CRF layer. Specifically, Chinese characters and dictionary features are first projected into dense vector representations, then they are fed into the convolutional layers to capture contextual features. Finally, a CRF layer is employed to capture dependencies between neighboring tags.

\subsection{Embedding Layer}

Given a clinical sentence $X={[x]}_1^n$, which is a sequence of $T$ characters, the first step is to map discrete language symbols, including the characters and their corresponding dictionary features, to dense embedding vectors. Formally, we first look up character embedding ${\bm{x}}_i \in \mathbb{R}^{d_x}$ from character embedding matrix $W_x$ for each character $x_i$, where $i \in \{1,2,\ldots,n\}$ indicates $x_i$ is the $i$-th character in $X$, and ${d_x}$ is a hyper-parameter indicating the size of character embedding. We also look up dictionary feature embedding ${\bm{d}}_i \in \mathbb{R}^{d_d}$ from dictionary feature embedding matrix $W_d$ for each dictionary feature which $x_i$ belongs to, where ${d_d}$ is a hyper-parameter indicating the size of dictionary feature embedding. The final embedding vector is created by concatenating ${\bm{x}}_i$ and ${\bm{d}}_i$ as ${\bm{e}_i}={\bm{x}}_i \oplus {\bm{d}}_i$, where $\oplus$ is the concatenation operator. It can be seen as an ensemble of a knowledge-based dictionary method and a data-driven deep learning method.

Specifically, as for the dictionary features, given a sentence $X$ and an external dictionary $D$, we first use the classic Bi-Directional Maximum Matching (BDMM) algorithm~\cite{gai2014} to segment $X$. Then, each character $x_i$ is labeled as the type of the entity which $x_i$ belongs to, as shown in the second line of Table \ref{table:labelingFormat}. Note that the dictionary features also take the position of a character in an entity into account via the BIEOS tag scheme. More details can be seen in our previous work~\cite{wang2018incorporating}.

\subsection{Convolutional Layer}

The convolutional layers used in the BDCNN-CRF model consists of two separate parts. The left part has two residual blocks with different dilation factors. The right part is a standard convolutional layer with batch normalization~\cite{ioffe2015batch}. The final output of the convolutional layers is the sum of the separate output of the two parts.

\begin{figure*}[!ht]
\begin{center}
\subfigure[2-layer Standard Convolutions]{
\label{fig:StandardConv} 
\includegraphics[height=1.2in]{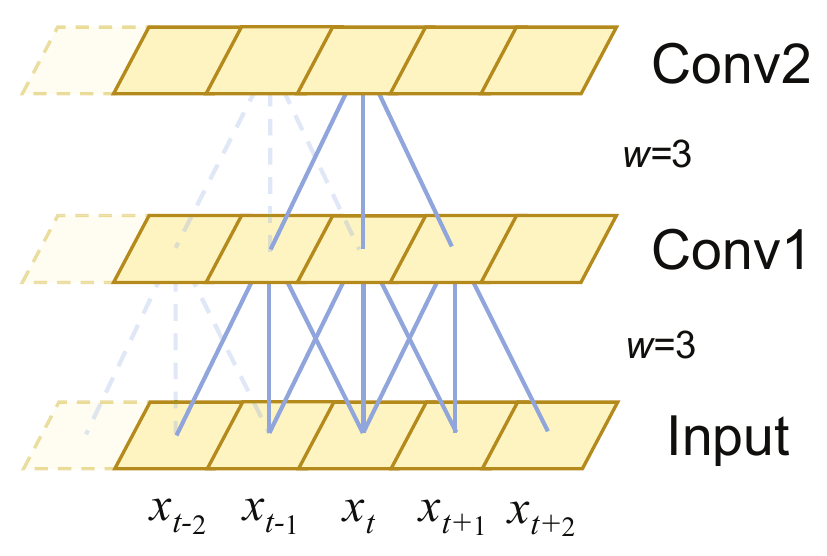}}
\hspace{0.5in}
\subfigure[2-layer Dilated Convolutions]{
\label{fig:DilatedConv} 
\includegraphics[height=1.2in]{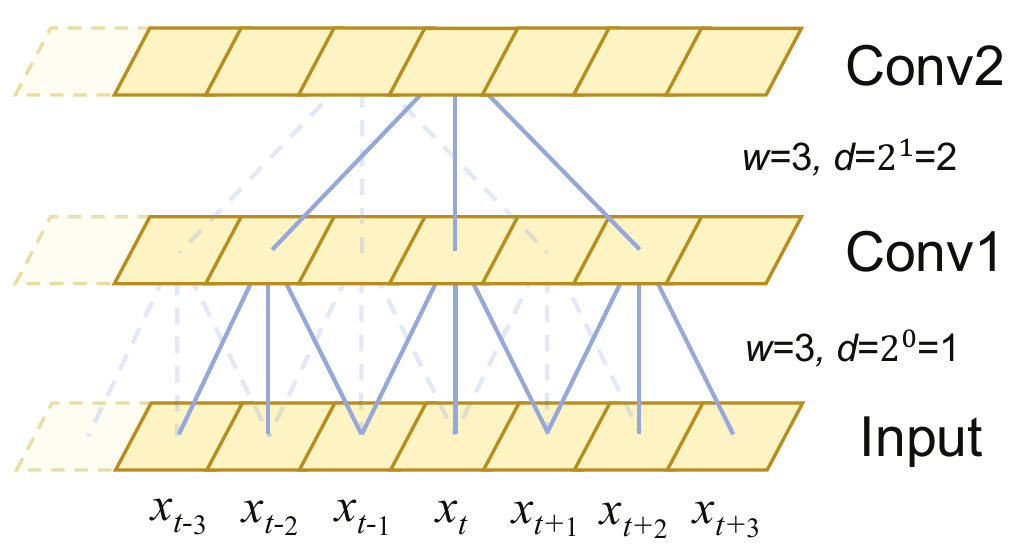}}
\hspace{0.5in}
\subfigure[Residual Block]{
\label{fig:ResidualBlock} 
\includegraphics[height=1.5in]{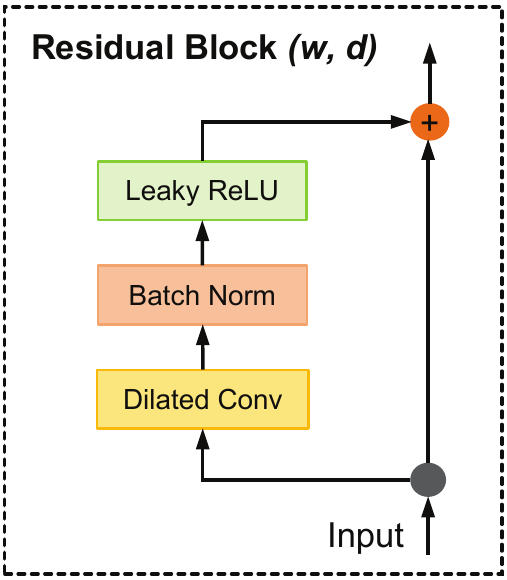}}
\caption{Examples of the convolutional layers.}
\label{fig:subfig} 
\end{center}
\end{figure*}

\subsubsection{Standard Convolution}

The standard convolutions~\cite{kalchbrenner2014convolutional} have been widely used in natural language processing tasks~\cite{Nguyen2015Relation,Dong2015Question,hsu2017hybrid}. Given a widow size $w=2l+1$, a filter is seen as a weight matrix $\bm{f} = [\bm{f}_{-l}, \bm{f}_{-l+1}, \ldots, \bm{f}_l]$ ($\bm{f}_i$ is a column vector of size $d_x$ + $d_d$)\footnote{If the widow size is even, it is similar to the odd one that $w=2l$ and $\bm{f} = [\bm{f}_{-l+1}, \bm{f}_{-l+1}, \ldots, \bm{f}_{l}]$.}. The core of the standard convolutional layer is obtained from the application of the convolutional operator on the two matrices $\bm{X}$ and $\bm{f}$ to produce a feature sequence $\bm{s} = [s_1, s_2, \ldots, s_n]$:
\begin{equation}
    s_i = \sum_{j=-l}^{l} \bm{f}_{j} \cdot \bm{x}_{i+j} + b
\end{equation}
where $b$ is a bias term and zero padding is conducted that tokens outside the input sequence $\bm{X}={[\bm{x}]}_1^n$ will be treated as zeros. This process can then be replicated for various filters with the same window size to capture different n-gram feature sequences of the input sentence. Suppose there are $f$ filters. The final output of the standard convolution $[\bm{y}]_1^n$ is created by concatenating all the n-gram features $\bm{s}_1, \bm{s}_2, \ldots, \bm{s}_f$ as ${\bm{y}_i}={\bm{s}}_{1,i} \oplus {\bm{s}}_{2,i} \oplus \ldots \oplus {\bm{s}}_{f,i}$, where $\oplus$ is the concatenation operator and ${\bm{s}}_{j,i}$ denotes the $i$-th element in $\bm{s}_j$. We also perform batch normalization~\cite{ioffe2015batch} after convolutions to accelerate training and avoid over-fitting.

\subsubsection{Dilated Convolution}

Given a convolutional filter $\bm{f} = [\bm{f}_{-l}, \bm{f}_{-l+1}, \ldots, \bm{f}_l]$ of a widow size $w=2l+1$ and the input sequence $\bm{X}={[\bm{x}]}_1^n$, a $d$ dilated convolution of $\bm{X}$ with respect the filter $\bm{f}$ can be described as:
\begin{equation}
    s_i = \sum_{j=-l}^{l} \bm{f}_{j} \cdot \bm{x}_{i+j*d} + b
\end{equation}
where $i \in  \{1, 2, \ldots, n\}$, $d$ is an exponential
dilation and $b$ is a bias term. Here we adopt zero padding, so tokens outside the sequence will be treated as zeros. Unlike the standard convolutions (i.e. $d$ = 1) that convolve each contiguous subsequence of the input sequence with the filter, a dilated convolution uses every $d$-th element in the sequence, but shifting the input by one at a time.

Repeated dilated convolution~\cite{yu2015multi} with an exponential dilation $d = d_b^{i-1}$ for layer $i$ in the network increases the receptive region of convolutional outputs exponentially with respect to the network depth, which results in drastically shortened computation paths compared with standard convolutions, as shown in Fig.~\ref{fig:StandardConv} and Fig.~\ref{fig:DilatedConv}.

\subsubsection{Residual Connection}

A residual block~\cite{He2015Deep} contains a branch leading out to a series of transformations $\mathcal{F}$, whose outputs are added to the input $x$ of the block:
\begin{equation}
    \bm{o} = \bm{x} + \mathcal{F}(\bm{x})
\end{equation}
Since the receptive field of a repeated dilated convolution depends on the network depth $D$ as well as filter size $w$ and dilation factor $d$, stabilization of deeper and larger dilated convolutional neural networks becomes important. The residual connection can effectively allow layers to learn modifications to the identity mapping rather than the entire transformation, which has repeatedly been shown to benefit very deep networks.

The residual block used in our model is shown in Fig.~\ref{fig:ResidualBlock}. Within a residual block, there exists two layers of dilated causal convolution and non-linearity, for which we utilize the leaky rectified linear unit (Leaky ReLU)~\cite{maas2013rectifier}. For normalization, we apply batch normalization~\cite{ioffe2015batch} to the convolutional filters.

\subsection{CRF layer}

For the character-based Chinese CNER task, it is beneficial to consider the dependencies of adjacent tags. For example, a B (begin) tag should be followed by an I (inside) tag or E (end) tag, and an I tag cannot be followed by a B tag or S (single) tag. Therefore, instead of making tagging decisions using the output of the convolutional layers independently, we employ a Conditional Random Field (CRF)~\cite{Lafferty2001Conditional} to model the tag sequence jointly and predict the CNER sequential tag result.

Generally, the CRF layer is represented by lines which connect consecutive output layers, and has a state transition matrix as parameters. With such a layer, we can efficiently use past and future tags to predict the current tag. We consider the matrix of scores $f_{\theta}([x]_{1}^{T})$ as the output of the convolutional layers. The element $[f_{\theta}]_{i,t}$ of the matrix is the score output by the network with parameters $\theta$, for the sentence $[x]_{1}^{T}$ and for the $i$-th tag, at the $t$-th character.  We introduce a transition score $[A]_{i,j}$ to model the transition from $i$-th state to $j$-th for a pair of consecutive time steps.  Note that this transition matrix is position independent. We now denote the new parameters for our whole network as $\tilde{\theta}=\theta\cup\{[A]_{i,j}\forall i,j\}$. The score of $[x]_{1}^{T}$ along with a path of tags $[i]_{1}^{T}$ is then given by the sum of transition scores and BDCNN network scores:
\begin{align}
    S([x]_{1}^{T},[i]_{1}^{T},\tilde{\theta})=\sum_{t=1}^T ([A]_{[i]_{t-1},[i]_{t}}+[f_{\theta}]_{[i]_{t},t})
\end{align}

The conditional probability $p([y]_{1}^{T}|[x]_{1}^{T},\tilde{\theta})$ is calculated with a softmax function:
\begin{align}
    p([y]_{1}^{T}|[x]_{1}^{T},\tilde{\theta})=\frac{e^{S([x]_{1}^{T},[y]_{1}^{T},\tilde{\theta})}}{\sum_{j} e^{S([x]_{1}^{T},[j]_{1}^{T},\tilde{\theta})}}
\end{align}
where $[y]_{1}^{T}$ is the true tag sequence and $[j]_{1}^{T}$ is the set of all possible output tag sequences.

The maximum conditional likelihood estimation is employed to train the model:
\begin{eqnarray}
    \log \! p([y]_{1}^{T}|[x]_{1}^{T}\! ,\tilde{\theta}) \! = \! S([x]_{1}^{T} \! ,[y]_{1}^{T} \!,\tilde{\theta}) \! - \! \log \! \sum_{\forall [j]_{1}^{T}} \! e^{S([x]_{1}^{T} \! ,[j]_{1}^{T} \!,\tilde{\theta})}
\end{eqnarray}

In our model, a dynamic programming algorithm~\cite{Rabiner1990A} is used to efficiently compute $[A]_{i,j}$ and the optimal tag sequences for inference.

\section{Experimental Studies}
\label{Sec:Experimental Studies}

In this section, we compare the proposed CD-RNN-CRF approach with state-of-the-art CNER methods. The best experimental results in tables are in bold.

\subsection{Dataset and Evaluation Metrics}
\label{SubSec:Dataset and Evaluation Metrics}

We use the CCKS-2017 Task 2 benchmark dataset\footnote{It is publickly available at \url{http://www.ccks2017.com/en/index.php/sharedtask/}} to conduct our experiments. This dataset contains 1,596 annotated instances (10,024 sentences) with five types of clinical named entities, including diseases, symptoms, exams, treatments and body parts. These annotated instances have been partitioned into 1,198 training instances (7,906 sentences) and 398 test instances (2,118 sentences). Each instance has one or several sentences. We further split these sentences into clauses by commas. Detailed statistics of different types of entities are listed in Table \ref{table:dataset}.

\begin{table}[!ht]
\begin{center}
\renewcommand\arraystretch{1.25}
\addtolength{\tabcolsep}{6pt}
\caption{Statistics of Different Types of Entities}
\label{table:dataset}
\begin{tabular}{|c|c|c|}
\hline
Type& Training Set & Test Set \\
\hline
Disease & 722 & 553\\
Symptom & 7,831 & 2,311 \\
Exam & 9,546 & 3,143 \\
Treatment & 1,048 & 465 \\
Body Part & 10,719 & 3,021 \\
\hline
Sum & 29,866 & 9,493 \\
\hline
\end{tabular}
\end{center}
\end{table}

In the following experiments, widely-used performance measures such as precision, recall, and F$_1$-score \cite{liu2014strategy,Zhou2014CORRELATION} are used to evaluate the methods.

\subsection{Experimental Settings}

The dictionary used in the experiments is constructed according to the lists of charging items and drug information in Shanghai Shuguang Hospital as well as some medical literature such as \begin{CJK*}{UTF8}{gbsn}《人体解剖学名词（第二版）》\end{CJK*} (\emph{Chinese Terms in Human Anatomy [Second Edition]}).

Parameter configuration may influence the performance of a deep neural network. The parameter configurations of the proposed approach are shown in Table~\ref{Tab:Parameter Configuration}. Note that the parameter selection of dilated convolutions are determined based on the experimental results in Section~\ref{Sec:Discussion and Analysis}.

To implement deep neural network models, we utilize the Keras library\footnote{\url{https://github.com/keras-team/keras}} with TensorFlow~\cite{abadi2016tensorflow} backend, and each model is run in a single NVIDIA GeForce GTX 1080 Ti GPU. Character embeddings and feature embeddings are initialized via word2vec~\cite{mikolov2013efficient} on both training data and test data. The models are trained by Adam optimization algorithm~\cite{kingma2014adam} whose parameters are same as the default settings. The best hyper-parameters are selected with grid search mechanism.

\begin{table}[!ht]
\begin{center}
\renewcommand\arraystretch{1.25}
\addtolength{\tabcolsep}{3pt}
\caption{Parameter Configurations of the Proposed Approach}
\label{Tab:Parameter Configuration}
\begin{tabular}{|l|l|}
\hline
\multicolumn{1}{|c|}{Parameter}  & \multicolumn{1}{|c|}{Value} \\
\hline
Size of character embedding  & $d_x$ = 128 \\
Size of feature embedding    & $d_d$ = 128 \\
Number of residual block                    & $n_r$ = 2 \\
Number of filters per residual block        & $f_d$ = 256 \\
Window size of dilated convolution          & $w_d$ = 2 \\
Dilation factor of the $i$-th residual block    & $d$ = $3^{i-1}$ \\
Number of filters for standard convolution  & $f_s$ = 256 \\
Window size of standard convolution         & $w_s$ = 3 \\
Batch size in training                      & $b$ = 128 \\
\hline
\end{tabular}
\end{center}
\end{table}

\subsection{Compared with State-of-the-art Models}

\subsubsection{Compared with Basic Bi-LSTM-CRF}

Since Gridach~\cite{Gridach2017Character}, Habibi et al.~\cite{Habibi2017Deep} and Zeng et al.~\cite{Zeng2017LSTM} have successfully employed Bi-LSTM-CRF models with no additional features for the English CNER. In this section, we first compare the proposed RD-CNN-CRF model with the Bi-LSTM-CRF models. As our previous work pointed that characters are better than words~\cite{Xia2017Clinical}. In this experiment, we adapt Bi-LSTM-CRF for the Chinese texts by using the Chinese characters instead of words as inputs. We implement two models with and without dictionary features. The hidden vector size of basic Bi-LSTM-CRF model is 256 which has the best performance through our test.
All comparative results are summarized in Table~\ref{table:ComparedWithSingleModels}.

\begin{table}[!ht]
\begin{center}
\renewcommand\arraystretch{1.25}
\addtolength{\tabcolsep}{1pt}
\caption{Comparative Results between Basic Bi-LSTM-CRF and Our RD-CNN-CRF Model}
\label{table:ComparedWithSingleModels}
\begin{tabular}{|c|c|c|c|}
\hline
        Methods          & Precision       &  Recall      &  F$_1$-score \\
\hline
Bi-LSTM-CRF (No Dict)    &  88.22  &  88.53  &  88.38 \\
Our RD-CNN-CRF (No Dict)      &  88.64  &  88.38  &  88.51 \\
\hline
Bi-LSTM-CRF (With Dict)  &  $\textbf{90.83}$  &  91.64  &  91.24 \\
Our RD-CNN-CRF (With Dict)    &  90.63  &  $\textbf{92.02}$  &  $\textbf{91.32}$ \\
\hline
\end{tabular}
\end{center}
\end{table}

From the table, we can observe that our RD-CNN-CRF model with dictionary features achieves the best performance, with a precision of $90.63\%$, a recall of $92.02\%$ and a F$_1$-score of $91.32\%$. In addition, the dictionary features can bring benefit, with an improvement of $2.61\%$ in terms of precision, $3.11\%$ in terms of recall and $2.86\%$ in terms of F$_1$-score for the basic Bi-LSTM-CRF. We can obtain the same observation on our RD-CNN-CRF model, i.e., an improvement of $1.99\%$ in terms of precision, $3.64\%$ in terms of recall and $2.81\%$ in terms of F$_1$-score.

To investigate training speed, we further compare our RD-CNN-CRF with the basic Bi-LSTM-CRF in F$_1$-score on the test set under different training time. Note that both models exploit dictionary features as inputs.

\begin{figure}[!htbp]
\begin{center}
\includegraphics[width=2.6in]{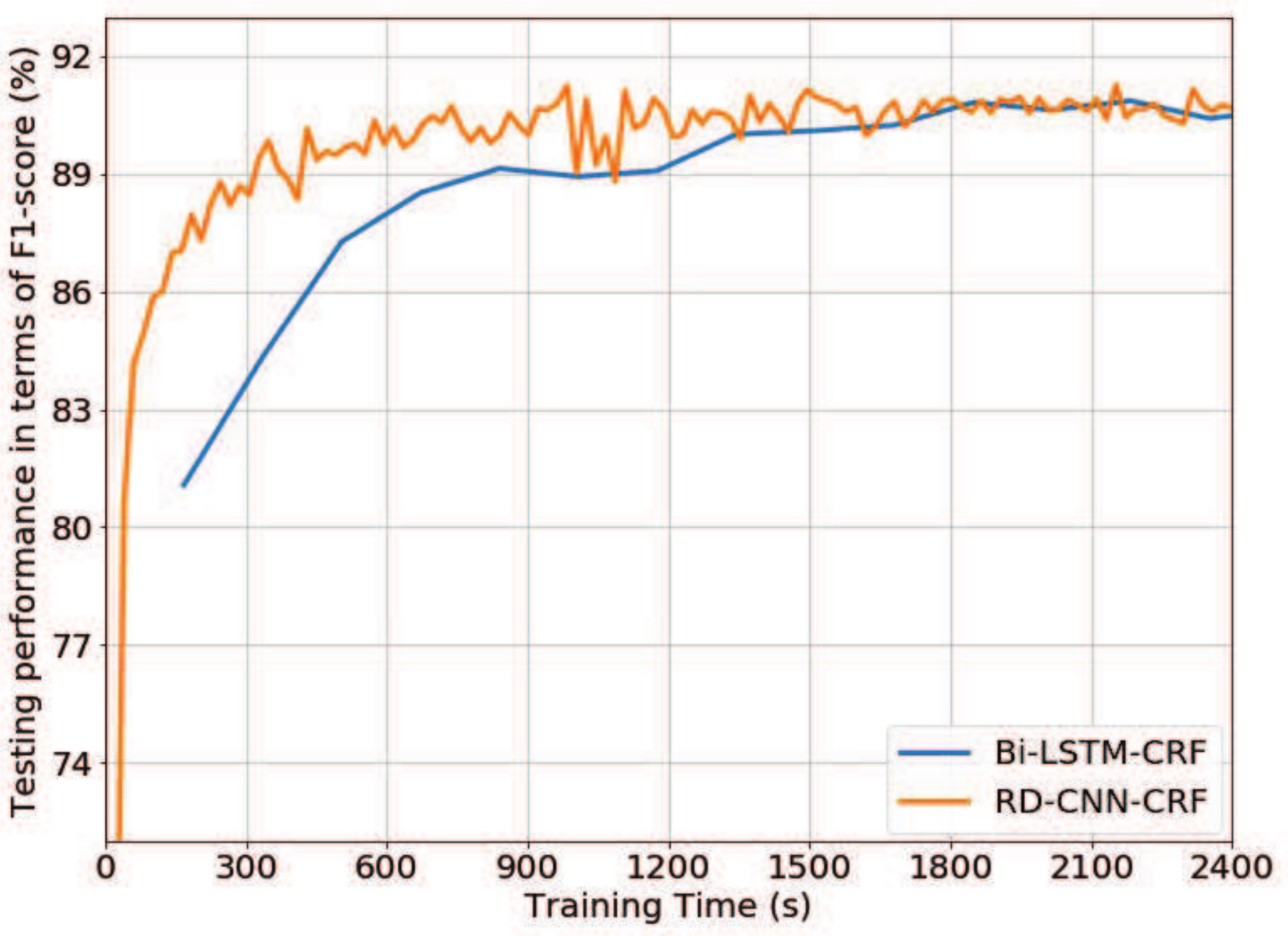}
\caption{Performance in terms of F$_1$-score with different training time.}
\label{fig:time}
\end{center}
\end{figure}

As shown in Fig.~\ref{fig:time}, our RD-CNN-CRF model begins to converge at about 900 second, while the basic Bi-LSTM-CRF model does not converge until 1,800 second. The training time of the basic Bi-LSTM-CRF model is about twice than our model. It is reasonable because LSTMs are dedicated sequence models which maintain a vector of hidden activations that are propagated through time, while convolution operations can perform in parallel which require less calculation time.

\subsubsection{Compared with Ensemble Models}

Besides the basic Bi-LSTM-CRF model, some existing methods attempt to incorporate additional features or additional approaches into Bi-LSTM-CRF models. They are known as ensemble models. For example, Li et al.~\cite{Li2017Recurrent} saw the Chinese CNER task as a sequence labeling problem in word level, and exploited a Bi-LSTM-CRF model to solve it. To improve recognition, They also used health domain datasets to create richer, more specialized word embeddings, and utilized external health domain lexicons to help word segmentation. Ouyang et al.~\cite{Ouyang2017Exploring} adopted Bi-RNN-CRF architecture with concatenated n-gram character representation to recognize Chinese clinical named entities. They also incorporated word segmentation results, part-of-speech tagging and medical vocabulary as features into their model. Xia and Wang~\cite{Xia2017Clinical} employed Bi-LSTM-CRF models with self-taught learning and active learning for CNER. Ensemble learning is also exploited to obtain the best recognition performance for all five types of clinical named entities. Hu et al.~\cite{Hu2017Compilation} developed a hybrid system based on rules, CRF and LSTM methods for the CNER task. They also utilized a self-training algorithm on extra unlabeled clinical texts to improve recognition performance. Note that except Li et al.~\cite{Li2017Recurrent}, the other systems all regarded the CNER task as a character level sequence labeling problem.

\begin{table}[!ht]
\begin{center}
\renewcommand\arraystretch{1.25}
\addtolength{\tabcolsep}{2pt}
\caption{Comparative Results between Ensemble Models and Our RD-CNN-CRF Model}
\label{table:comparisonOthers}
\begin{tabular}{|c|c|c|c|}
\hline
     Methods         & Precision     & Recall     & F$_1$-score \\
\hline
Li et al. \cite{Li2017Recurrent}     & -     & -     & 87.95 \\
Ouyang et al. \cite{Ouyang2017Exploring} & -     & -     & 88.85 \\
Xia and Wang \cite{Xia2017Clinical} *    & -     & -     & 89.88 \\
Hu et al. \cite{Hu2017Compilation}     & \textbf{94.49} & 87.79 & 91.02 \\
Hu et al. \cite{Hu2017Compilation} *    & 92.99 & 89.25 & 91.08 \\
Our RD-CNN-CRF & 90.63  &  $\textbf{92.02}$  &  $\textbf{91.32}$ \\
\hline
\multicolumn{4}{l}{\begin{tabular}[c]{@{}l@{}}\begin{minipage}{5.9cm}\vspace{1mm}\tiny \item[*] The results are obtained by allowing the use of external resources for self-training.\end{minipage} \end{tabular}}
\end{tabular}
\end{center}
\end{table}

The comparative results are shown in Table~\ref{table:comparisonOthers}. From the table, we observe that our proposed RD-CNN-CRF model achieves the best results among all the models. Li et al.~\cite{Li2017Recurrent} gets the worst performance because of two reasons. One is that their word-level approach inevitably has wrong segmentation, which leads to boundary errors when recognition. The other one is that since clinical texts are labeled in word level, there exists much more words than characters, so the corpus may contain many rare words which are difficult to be recognized. In fact, except for Li et al.~\cite{Li2017Recurrent}, the other methods all regarded the CNER task as a sequence labeling problem in character level. It shows the benefits of the character-level CNER for Chinese.

As for the rest character-level CNER approaches, Hu et al.~\cite{Hu2017Compilation} utilized an ensemble method which consists of three separated models, and finally gets $91.08\%$ in terms of F$_1$-score. It is the best one among the previous models, but is complex for practice. While we only exploit one model, and achieve an improvement $0.24\%$ in terms of F$_1$-score compared with Hu et al.~\cite{Hu2017Compilation}.

\section{Discussion and analysis}
\label{Sec:Discussion and Analysis}

In this section, we perform additional experiments to gain some understanding of our proposed RD-CNN-CRF model. Specifically, we conduct two groups of experiments to respectively investigate usefulness of model components and impact of different residual dilated convolutions.

\subsection{Usefulness of the Model Components}

To investigate the usefulness of model components, we study the effect of the combination between standard convolutions and residual dilated convolutions, and the interest of residual connections respectively.

\subsubsection{Effect of the Combination Between Two Types of Convolutions}

To study the effect of the combination between standard convolutions and residual dilated convolutions, we compare the proposed RD-CNN-CRF model with four alternative models. These alternative models are obtained by only using one type of convolution (i.e. only preserve one side of convolutional layers) or replace one type of convolution with the others (i.e. both sides of convolutional layers are the same).

\begin{table}[!ht]
\begin{center}
\renewcommand\arraystretch{1.25}
\addtolength{\tabcolsep}{2pt}
\caption{Comparative Results Between Different Convolutions}
\label{table:comparisonConvolutions}
\begin{tabular}{|c|c|c|c|}
\hline
     Methods         & Precision     & Recall     & F$_1$-score \\
\hline
Standard Conv Only     & 90.55  &  91.77  &  91.16 \\
Residual Blocks Only   & 90.35  &  91.91  &  91.12 \\
\hline
Both Standard Conv     & 90.28  &  92.18  &  91.21 \\
Both Residual Blocks   & 90.07  &  \textbf{92.45}  &  91.24 \\
\hline
RD-CNN-CRF    & \textbf{90.63}  &  92.02  &  \textbf{91.32} \\
\hline
\end{tabular}
\end{center}
\end{table}

Comparative results are presented in Table~\ref{table:comparisonConvolutions}. From this table, we observe that each half of the convolutional layers can perform well independently (F$_1$-score $\geq 91.12\%$). Moreover, we can obtain a better F$_1$-score when they combines together.

\subsubsection{Interest of Residual Connections}

To analyze the interest of residual connections, we compare the performance of models with or without residual connections. Comparative results are summarized in Table \ref{table:comparisonResidualConnections}.

\begin{table}[!ht]
\begin{center}
\renewcommand\arraystretch{1.25}
\addtolength{\tabcolsep}{2pt}
\caption{Comparative Results of Models With and Without Residual Connections}
\label{table:comparisonResidualConnections}
\begin{tabular}{|c|c|c|c|}
\hline
     Methods         & Precision     & Recall     & F$_1$-score \\
\hline
No Residual Connections  & 89.85  &  \textbf{92.21}  &  91.02 \\
With Residual Connections     & \textbf{90.63}  &  92.02  &  \textbf{91.32} \\
\hline
\end{tabular}
\end{center}
\end{table}

As shown in Table~\ref{table:comparisonResidualConnections}, the model with residual connections outperforms that without residual connections. The benefit brought by residual connections is $0.30\%$ in terms of F$_1$-score. It is reasonable because residual connections can help the model to utilize both semantic (i.e. high-level) and low-level features.

\subsection{Impact of Different Residual Dilated Convolutions}

In this section, we experimentally investigate the influence of several important parameters, namely the block number $n_r$, dilation factor $d$, filter number $f_d$ and window size $w_d$.

\subsubsection{Impact of Different Block Number and Dilation Factor}

Contextual information is very useful for CNER. Once the window size is determined, there are two main approaches to capture more contextual information. The one is to make the network deeper by increasing the block number $n_r$. The other one is to expand the receptive field by increasing the dilation factor $d$. To investigate the impact of the block number $n_r$ and dilation factor $d = d_b^{i-1}$, we compare the performance of models with different parameter values. Table~\ref{table:comparisonDeepDilation} displays the comparative results.

\begin{table}[!ht]
\begin{center}
\renewcommand\arraystretch{1.25}
\addtolength{\tabcolsep}{2pt}
\caption{Comparative Results (F$_1$-score) of Models With Different Block Number $n_r$ and Dilation Factor $d = d_b^{i-1}$}
\label{table:comparisonDeepDilation}
\begin{tabular}{|c|c|c|c|}
\hline
\multirow{2}{*}{\begin{tabular}[c]{@{}c@{}}Block\\ Number\end{tabular}} & \multicolumn{3}{c|}{Dilation Factor} \\ \cline{2-4}
      & $d_b$ = 1    & $d_b$ = 2    & $d_b$ = 3   \\
\hline
$n_r$ = 2  & 91.21  & 91.21  & \textbf{91.32} \\
\hline
$n_r$ = 3  & 91.30  & 91.29  & 90.92 \\
\hline
$n_r$ = 4  & 91.12  & 90.77  & 90.45 \\
\hline
\end{tabular}
\end{center}
\end{table}

From the Table~\ref{table:comparisonDeepDilation}, we are able to obtain several interesting observations. Firstly, when $n_r = 2$ and $d_b = 3$, the model achieves the best performance with a F$_1$-score of $91.32\%$. This confirms the benefit of expanding the receptive field. Secondly, when $n_r = 3$ and $d_b = 1$, the F$_1$-score is $91.30\%$, which is only slightly lower than $91.32\%$. It indicates that making the network deeper has a same effect as well as expanding the receptive field. Thirdly, when $n_r = 4$, performance of the model gradually deteriorates as the increase of $d_b$. We can obtain same observation when $d_b = 3$ and $n_r$ gradually increases. The above observations show that long-distance contextual information is not necessary to the Chinese CNER task.

\subsubsection{Influence of Different Filter Number and Window Size}

We further investigate the influence of different numbers and different window sizes of dilated filters by running the proposed RD-CNN-CRF model on filter numbers of 128, 256, 384 and 512, and window sizes of 1, 2, 3 and 4. Table~\ref{table:comparisonFilterWindow} presents the model performance on F$_1$-score. Note that due to the residual connections, the concatenated embedding size is set the same as the filter number here.

\begin{table}[!ht]
\begin{center}
\renewcommand\arraystretch{1.25}
\addtolength{\tabcolsep}{2pt}
\caption{Comparative Results (F$_1$-score) of Models With Different Filter Number $f_d$ and Window Size $w_d$}
\label{table:comparisonFilterWindow}
\begin{tabular}{|c|c|c|c|c|}
\hline
\multirow{2}{*}{\begin{tabular}[c]{@{}c@{}}Filter\\ Number\end{tabular}} & \multicolumn{4}{c|}{Window Size}       \\ \cline{2-5}
  & $w_d$ = 1  & $w_d$ = 2  & $w_d$ = 3  & $w_d$ = 4     \\
\hline
$f_d$ = 128     & 91.09      & 91.29      & 90.96      & 90.74 \\
\hline
$f_d$ = 256     & 91.19      & \textbf{91.32}   & 90.93      & 90.77 \\
\hline
$f_d$ = 384     & 91.20      & 90.92      & 90.63      & 90.91 \\
\hline
$f_d$ = 512     & 91.17      & 91.08      & 90.92      & 90.67 \\
\hline
\end{tabular}
\end{center}
\end{table}

Generally speaking, the larger window size can bring more contextual information for CNER. Except the last line, the F$_1$-score first grows then drops down with the increase of the window size, and the window size of 2 outperforms the other settings. It indicates that the window size is not required to be large since the adoption of the dilated mechanism. And with filter number f$_d$ = 256 and window size w$_d$ = 2, our model achieves the best performance.

\section{Conclusion}
\label{Sec:Conclusion and Future Work}

In this paper, we propose a Residual Dilated Convolutional Neural Network with Conditional Random Field (RD-CNN-CRF) for the Clinical Named Entity Recognition (CNER). Character-level sequence labeling task is conducted to avoid introducing noise caused by segmentation errors in Chinese. Dictionary features are also utilized to help recognize rare and unseen clinical named entities. Unlike existing RNN-based models, we first introduce dilated convolutions to capture the contextual information, and employ residual connections to utilize both semantic (i.e. high-level) and low-level features. Finally, a CRF is used as the output interface to obtain the optimal sequence of tags over the entire sentence. Experimental results on the CCKS-2017 Task 2 benchmark dataset demonstrate that our proposed RD-CNN-CRF method achieves a highly competitive performance compared to state-of-the-art RNN-based methods, i.e., reaching $90.63\%$, $92.02\%$ and $91.32\%$ in terms of precision, recall and F$_1$-score, respectively.

\section*{Acknowledgment}

We would like to thank the referees for their useful comments and suggestions. We also want to thank Yichao Yin (Shanghai Shuguang Hospital) who has helped us very much in data fetching and data cleansing. This work is supported by the National Key R\&D Program of China for ``Precision Medical Research" (No.~2018YFC0910500), the National Natural Science Foundation of China (No.~61772201) and National Major Scientific and Technological Special Project for ``Significant New Drugs Development" (No.~2018ZX09201008).

\bibliographystyle{IEEEtran}
\bibliography{IEEEexample}

\end{document}